\documentclass[10pt,twocolumn,letterpaper]{article}

\usepackage[pagenumbers]{cvpr} 

\usepackage{stfloats}
\usepackage{pifont}
\usepackage{makecell}
\usepackage{color}
\usepackage{bbding}
\usepackage{graphicx}
\usepackage{epstopdf}
\usepackage{amsmath}
\usepackage{amssymb}
\usepackage{booktabs}
\usepackage[mathscr]{eucal}
\usepackage{dutchcal}
\usepackage{algorithm}
\usepackage{algorithmicx}
\usepackage{algpseudocode}
\usepackage{multirow}
\floatname{algorithm}{Algorithm} 

\usepackage[pagebackref,breaklinks,colorlinks]{hyperref}

\usepackage[capitalize]{cleveref}
\crefname{section}{Sec.}{Secs.}
\Crefname{section}{Section}{Sections}
\Crefname{table}{Table}{Tables}
\crefname{table}{Tab.}{Tabs.}

\def\cvprPaperID{8826} 
\def\confName{CVPR}
\def\confYear{2023}

\usepackage{xcolor}

\begin{document}

\title{Transferability Estimation Based On Principal Gradient Expectation}


\author{
Huiyan Qi\textsuperscript{\rm 1}, Lechao Cheng\textsuperscript{\rm 2}, Jingjing Chen\textsuperscript{\rm 1}, Yue Yu\textsuperscript{\rm 1}, Xue Song\textsuperscript{\rm 1}, Zunlei Fengg\textsuperscript{\rm 3}, Yu-Gang Jiang\textsuperscript{\rm 1}\\
\textsuperscript{\rm 1}School of Computer Science \& Shanghai Collaborative Innovation Center of Intelligent Visual Computing,\\
Fudan University\\
\textsuperscript{\rm 2}Zhejiang Lab\\
\textsuperscript{\rm 3}Zhejiang University\\
}

\maketitle

\begin{abstract}
   Transfer learning aims to improve the performance of target tasks by transferring knowledge acquired in source tasks. The standard approach is pre-training followed by fine-tuning or linear probing. Especially, selecting a proper source domain for a specific target domain under pre-defined tasks is crucial for improving efficiency and effectiveness. It is conventional to solve this problem via estimating transferability. However, existing methods can not reach a trade-off between performance and cost. To comprehensively evaluate estimation methods, we summarize three properties: stability, reliability and efficiency. Building upon them, we propose Principal Gradient Expectation (PGE), a simple yet effective method for assessing transferability. Specifically, we calculate the gradient over each weight unit multiple times with a restart scheme, and then we compute the expectation of all gradients. Finally, the transferability between the source and target is estimated by computing the gap of normalized principal gradients. Extensive experiments show that the proposed metric is superior to state-of-the-art methods on all properties.
\end{abstract}

\section{Introduction}
\label{sec:intro}

Traditional machine learning work for predicting unseen testing instances with the knowledge learned from training data~\cite{day2017survey, weiss2016survey}. Since there is always a distribution gap between training data and testing data, the inference performance is lower than expected~\cite{zhuang2020comprehensive, tan2018survey}. However, as obtaining appropriate training data is challenging, a surge of interest emerges in transferring knowledge from the source task into the target task based on the pre-training mechanism. In the realm of computer vision, the prevalent approaches are linear probing~\cite{chen2021empirical} and fine-tuning. Linear probing adapts the model outputs by learning a task-related head layer while fine-tuning adapts the parameters of the entire model. Both linear probing and fine-tuning leverage models pre-trained on large-scale datasets for downstream tasks. 
\begin{figure}[htbp]
  \centering
   \includegraphics[width=1\linewidth]{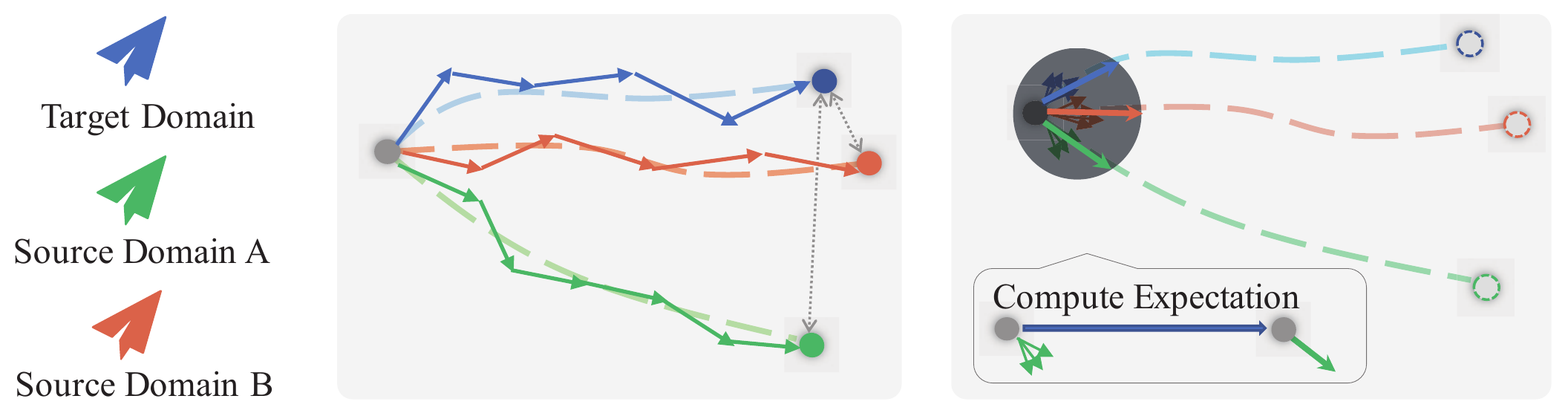}

   \caption{The figure on the left denotes that the model's parameters are updated by first-order gradients in the optimization process. Intuitively, the gap between the optimal points of the source and target domains can reflect transferability. However, the point of the target is invisible. To solve this problem, our PGE collects first-order gradients and computes the expectation of all gradients as on the right.}
   \label{fig:idea}
\end{figure}
Prominent models, including BERT~\cite{devlin2018bert} and GPT-3~\cite{brown2020language} for natural language processing, the ResNet~\cite{he2016deep} and vision  transformer~\cite{dosovitskiy2020image,liu2021Swin,liu2021swinv2} variants pre-trained on ImageNet~\cite{russakovsky2015imagenet} for vision tasks, and the CLIP~\cite{radford2021learning,ramesh2022hierarchical,saharia2022photorealistic} variants for multimodal learning, have significantly outperformed previous approaches in their fields of expertise. However, it is intuitive that a proper source domain is more crucial than the scale of pre-training data. Take CIFAR100~\cite{krizhevsky2009learning} for example, CIFAR10~\cite{krizhevsky2009learning} may be a better source than ImageNet~\cite{russakovsky2015imagenet}. Therefore, researchers are dedicated to designing transferability estimating algorithms to determine the best source for the target domain efficiently. 
\par
To estimate transferability, researchers have proposed several transferability metrics, including LEEP~\cite{nguyen2020leep}, LogMe~\cite{you2021logme}, H-score~\cite{bao2019information}, NCE~\cite{tran2019transferability}, and GBC~\cite{pandy2022transferability}. These metrics utilize the distribution in label and feature spaces of the target domain. 
However, we found that sampling from the target domain with certain approaches as previous works~\cite{nguyen2020leep, pandy2022transferability} could change the data distribution, which may lead to inferior results lacking in stability and accuracy. Furthermore, the calculations of mentioned metrics demand the distributions in feature spaces of pre-trained models. But the pre-training process is time-consuming, which is harmful to efficiency.
\par
As mentioned above, existing metrics can not reach a trade-off between performance and cost, leading to low usability. To comprehensively evaluate these estimation methods, we summarize three properties that a good transferability estimation should possess for the first time. We consider that a qualified measurement should be stable, reliable, and efficient. Stability implies that the transferability metric value would not change violently with a slight oscillation of data distribution. Reliability reflects the accuracy of selecting a proper source domain. Efficiency requires concise calculations and low time costs. 
\par
In this paper, we suggest a straightforward yet effective technique for transferability estimation that satisfies the three properties, dubbed Principal Gradient Expectation (PGE). Similarly to previous metric methods~\cite{nguyen2020leep, you2021logme, bao2019information,tran2019transferability, pandy2022transferability}, our approach applies to the scenario where both the source and target tasks are a single task. Specifically, we consider the optimization procedure to be an approximation between the initial and the optimal state in the parameter space. Moreover, the optimal state gap between the source and target domain indicates transferability. However, as shown in Figure~\ref{idea}, the optimal point of the target is invisible. To quantify the gap, we collect first-order gradients of the model’s backbone multiple times and compute the expectation of all gradients (on the right of Figure~\ref{idea}). The expectation is named principal gradient expectation, which guarantees stability. Then, we design a scheme to compute the gap between principal gradients. Specifically, we deflate the gap of the principal gradient expectations between the source and target by Schwarz inequality~\cite{steele2004cauchy} to achieve a lower bound as the quantitative transfer gap. In the calculation process of the proposed metric described above, the three properties are guaranteed. As we do not utilize the distribution of the target domain and collect gradients multiple times to compute the expectation of gradient, our PGE could produce stable results. Considering that supervised pre-training is more sensitive than unsupervised pre-training~\cite{zhao2020makes,islam2021broad}, we obtain the first-order gradient in an unsupervised mode. This operation facilitates the promotion of accuracy in selecting a proper source domain. In addition, the process of estimating transferability with principal gradient expectation does not depend on pre-training parameters, which helpfully reduces the time cost. 
In the experiments, we show that our proposed method outperforms state-of-the-art methods in terms of stability, reliability and efficiency. To conclude, the main contributions of this work can be summarized as follows: 
\begin{itemize}
    \item For the first time, we have summarized the properties that a qualified transferability metric should satisfy and proposed corresponding evaluation methods. The properties are stability, reliability and efficiency.
    \item We propose a novel approach (PGE) based on the principal gradient expectation for assessing transferability. Moreover, our method satisfied the above properties.
    \item We show through experiments that the proposed PGE is strongly related to the transferred performance. Our approach is simple yet effective and shows its superiority over other methods. 
\end{itemize}

\section{Related Work}\label{sec:related}
Our research is closely aligned with the techniques of network adaptation and transferability estimation. Consequently, we will now delve into the body of literature surrounding these fields.
\par
\subsection{Network Adaptation}
Network adaptation is prevalent in the computer vision field nowadays. This technique involves adapting a pre-trained neural network to suit a new task in situations where data is scarce. Linear probing (LP) and Fine-tuning (FT) are typical methods for network adaptation. In the training process of LP, the pre-trained parameters are frozen, while training the head layers associated with the task~\cite{belinkov-etal-2017-neural, peters-etal-2018-deep}. As for FT, the parameters of the entire network are fine-tuned for downstream tasks to improve transfer accuracy~\cite{hewitt-manning-2019-structural}. 
Researchers have also explored various methods to improve network adaptation. To alleviate overfitting, ~\cite{zhang2020side}combined a lightweight “side” network with pre-trained models. ~\cite{lee2020residual}integrated each layer of a model into the fine-tuned model. This method can balance the performance of the source and target without increasing the model size. ~\cite{cai2020tinytl}improved the ability of a model to extract features by learning residual feature maps and significantly lessened memory usage. 
Upstream bias mitigation(UBM) ~\cite{jin2020transferability} has been found to benefit fine-tuning language models. Sparse pruning~\cite{dettmers2019sparse}reduced the predictive error via allocating pruning weights according to the zero-value weights in each layer. ~\cite{gale2019state}evaluated three pruning methods and found simple method may yield better results. Meta-learning has been applied to network adaptation and training dynamic networks~\cite{li2020international}. Some studies~\cite{trivedi2019dyrep, santoro2016meta} have combined meta-learning with reinforcement learning to improve the performance on new tasks. Recent work~\cite{finn2017model} generalized given tasks to new tasks.
\par
In addition, Neural Architecture Search(NAS) has also received significant attention. Previous NAS works~\cite{zoph2018learning, tan2019mnasnet, liu2018darts} are based on reinforcement learning algorithms and their cost was high. ENAS~\cite{pham2018efficient} simplified the search process by sharing parameters but it led to lower accuracy. Different NAS works~\cite{liu2018darts, dong2019searching, fang2020densely} have attempted to reduce the search cost and improve accuracy in recent years. ~\cite{fang2020fna++} achieved good results by automatically adjusting the network architecture to adapt to new tasks. More recently, ~\cite{lester2021power, li2021prefix}utilized prompts which consist of task-specific vectors and optimized prompts via gradients. Different studies~\cite{ju2022prompting, ge2022domain, yao2021cpt} applied prompts to multimodal tasks. ~\cite{sohn2022visual}trained vision transformers with prompts and generative knowledge. In this work, we adopt LP and FT as the adaptation approaches to evaluate our algorithm by measuring the correlation between the adaptation results and the transferability computed with the principal gradient.

\subsection{Transferability Estimation}
To properly evaluate the transferability of pre-trained models, Negative Conditional Entropy (NCE)~\cite{tran2019transferability} utilizes a metric derived from information theory to measure the transferability and difficulty of classification tasks. NCE assumes that the images in the source domain and the target domain are the same, but their labels are distinct. They then calculate the transferability score with the negative conditional entropy between the target and the source labels.
Contemporary work H-score~\cite{bao2019information} leverages statistics and information theory to quantify the transferability of feature representations. However, H-score can only be employed in classification tasks. While LEEP~\cite{nguyen2020leep} focuses on computing the joint probability over pseudo labels and the target labels to yield the log expectation of the empirical predictor. But LEEP will obtain different transferability scores when models with identical feature extractors and distinct classification heads. LogMe~\cite{you2021logme} solve this issue by directly estimating the maximum value of label evidence given features extracted by pre-trained models. It models each target label as a linear model with Gaussian noise, and then optimizes the prior distribution parameters to obtain the average maximum (log) evidence of labels given the target instances’ embeddings. Recently, GBC~\cite{pandy2022transferability} desires to measure the amount of overlap between target classes in the feature space of the pre-trained model. While in this work, the gradients of the model's backbone are employed in our estimation instead of distribution.

\section{Method}\label{sec:method}

\subsection{Preliminaries}
Considering a deep model $\boldsymbol{\mathit M}(w,h)$, we denote the feature extractor as $w$ and the task-related layer as $h$. For $N$ alternative sources $\{(\mathcal D_{s(i)}, \mathcal T_{s(i)})\}_{i=1}^N$, $\mathcal D_{s(i)}$ and $\mathcal T_{s(i)}$ represent the source domain and the source task, respectively. The target is denoted as $(\mathcal D_t, \mathcal T_t)$, where $\mathcal D_t$ and $\mathcal T_t$ separately indicate the target domain and the target task.
The goal of this work is to determine the appropriate source via quantify the transferability between each source $(\mathcal D_{s(i)}, \mathcal T_{s(i)})$ and the target $(\mathcal D_t, \mathcal T_t)$, as selecting a proper source is critical to improving efficiency and effectiveness. Moreover, we use $\mathcal G[(\mathcal D_{s(i)}, \mathcal T_{s(i)});(\mathcal D_t, \mathcal T_t)]$ to denote the transfer gap from an available source to the target.

After analyzing the previous methods~\cite{tran2019transferability, bao2019information, nguyen2020leep, you2021logme, pandy2022transferability}, we recapitulate three characteristics of a good transferability metric: stability, reliability, and efficiency. Next, we provide a comprehensive explanation of these propositions.
\par
\textbf{Proposition 1 (Stability).} Let $\emph{I}$ denote a random sampling function (further details about the sampling methods are described in Sec.~\ref{sec:robust_eva}). $\emph{I}(\mathcal D_t)$ represents a subset drawn according to $\mathcal D_t$. The stability indicating the difference between $\mathcal G[(\mathcal D_{s(i)}, \mathcal T_{s(i)});(\boldsymbol{\mathcal D_t}, \mathcal T_t)]$ and $\mathcal G[(\mathcal D_{s(i)}, \mathcal T_{s(i)});(\boldsymbol{\emph{I}(\mathcal D_t)}, \mathcal T_t)]$ is bounded, which can be described as:
\begin{equation}
|\mathcal G[(\mathcal D_{s(i)}, \mathcal T_{s(i)});(\boldsymbol{\mathcal D_t}, \mathcal T_t)] - \mathcal G[(\mathcal D_{s(i)}, \mathcal T_{s(i)});(\boldsymbol{\emph{I}(\mathcal D_t)}, \mathcal T_t)]|\leq \epsilon.
\end{equation}
Stability is essential since a finite dataset is a sub-sample in the manifold space of the data distribution.
\par
\textbf{Proposition 2 (Reliability).} $\mathcal A[(\mathcal D_{s(i)}, \mathcal T_{s(i)});(\mathcal D_t, \mathcal T_t)]$ is defined as the performance of the target domain transferred from an available source. 
Moreover, a lower transfer gap indicates better transfer performance. Thus,
$\mathcal G[(\mathcal D_{s(i)}, \mathcal T_{s(i)});(\mathcal D_t, \mathcal T_t)] < \mathcal G[(\mathcal D_{s(j)}, \mathcal T_{s(j)});(\mathcal D_t, \mathcal T_t)]$ always enables $\mathcal A[(\mathcal D_{s(i)}, \mathcal T_{s(i)});(\mathcal D_t, \mathcal T_t)] > \mathcal A[(\mathcal D_{s(j)}, \mathcal T_{s(j)});(\mathcal D_t, \mathcal T_t)]$.


\textbf{Proposition 3 (Efficiency).} The estimation of the transferability from source to target is supposed to have low computation and complexity.
\par

Previous studies such as ~\cite{zhao2020makes, islam2021broad}, have demonstrated that the transferability is mainly related to the model's backbone. However, existing methods employ task-specific layers $h$ besides the model's backbone to calculate the transferability, which may harm reliability and lead to high costs. Our method only utilizes the gradient of the backbone to estimate the transferability. Furthermore, to reach a trade-off between computing efficiency and efficacy, we use the first-order approximation of the loss.

\subsection{Principal Gradient Expectation (PGE)}\label{sec:fast_approx}

We consider the optimization procedure to be a distance approximation between the initial and the optimal points in the parameter space. To balance efficiency and effectiveness, we adopt the first-order gradient expectation. Given a model $\boldsymbol{\mathit M}$ and a target $(\mathcal D_t, \mathcal T_t)$, $\mathcal L$ denotes the loss function. The model is initialized with random weights $\theta_0$ where $\theta_0 \sim \mathscr N(\mathbf{0}, I)$. We first compute the gradient of backbone parameters, denoted as $\nabla \mathcal L(\theta_0)$. Instead of updating model parameters $\theta$ with $\nabla \mathcal L(\theta_0)$, we then re-initialize the weights of the model and collect $\nabla \mathcal L(\theta_0)$ multiple times to compute the expectation of gradients. We believe gradients reflect the inherent characteristics of the source task (target task). Furthermore, re-initializing multiple times and computing the expectation reduces the impact of abnormal gradients. This expectation value is defined as Principal Gradient Expectation (PGE), which can be formulated as follows:
 

\begin{equation}
PGE = \mathbb E_{\theta_0}[\nabla \mathcal L(\theta_0)].
\end{equation}
By making use of the definition of PGE, we compute the PGE for source $(\mathcal D_{s(i)}, \mathcal T_{s(i)})$ and target $(\mathcal D_t, \mathcal T_t)$ with their own loss, respectively:
\begin{equation}
PGE_{(\mathcal D_{s(i)}, \mathcal T_{s(i)})}=\mathbb E_{\theta_0}[\nabla \mathcal L_{(\mathcal D_{s(i)}, \mathcal T_{s(i)})}(\theta_0)],
\end{equation}
\begin{equation}
PGE_{(\mathcal D_t, \mathcal T_t)}=\mathbb E_{\theta_0}[\nabla \mathcal L_{(\mathcal D_t, \mathcal T_t)}(\theta_0)].
\end{equation}
\subsection{Transferability Metric based on PGE}

In our transfer setting, the restriction for the source and target tasks is that the backbone of models must be the same. Moreover, we require the source task $\mathcal T_s$ and target task $\mathcal T_t$ to be a single task as in most existing works. And $\mathcal T_t$ is not necessary to be the same as $\mathcal T_s$. So computing transferability with PGE is task-irrelevant.

\par
We consider a situation in which the model $\boldsymbol{\mathit M}$ only has one single parameter $\theta$. The initialized value of $\theta$ is denoted as $\theta_0$ and the optimal value is denoted $\theta^*$. 
$\mathcal L(\theta_0)$ is the initial loss value at $\theta_0$ and 
$\mathcal L(\theta^*)$ is the loss value at $\theta^*$. $\mathcal L^{'}(\theta_0)$ represents the derivative at $\theta_0$. 
According to Taylor's formula, the loss value at $\theta^*$ could be expressed in the following form:
\begin{equation}
\label{eq:loss_taylor}
\mathcal L(\theta^*) = \mathcal L(\theta_0) + \mathcal L^{'}(\theta_0)(\theta^* - \theta_0) + R(\theta_0).
\end{equation}
$R(\theta_0)$ is the remainder beyond the first-order approximation. 
Then we rewrite the Eq.~\ref{eq:loss_taylor}.
\begin{equation}
\label{eq:loss_taylor_change}
(\theta^* - \theta_0)  + \frac{R(\theta_0)}{\mathcal L^{'}(\theta_0)} = \frac{\mathcal L(\theta^*) - \mathcal L(\theta_0)}{\mathcal L^{'}(\theta_0)} .
\end{equation}
In Eq.~\ref{eq:loss_taylor_change}, $(\theta^* - \theta_0)$ can be regarded as the optimization distance of $ \theta$ from the initial state to the optimal state. However, since the optimal parameter $\theta^*$ is always invisible in practice, we require a more simplified definition for the gap. The term $\frac{R(\theta_0)}{\mathcal L^{'}(\theta_0)}$ contains high-order derivatives, which implies the complexity of the optimization surface. We propose a hypothesis that the optimization distance and the optimization surface's complexity can reflect the optimization difficulty of $ \theta$, and they are proportional to $1/\mathcal L^{'}(\theta_0)$. Therefore, we define $1/\mathcal L^{'}(\theta_0)$ as a factor of the optimization difficulty from $\theta_0$ to $\theta^*$ as follows:



\begin{equation}
\label{eq:factor}
\mathcal{f}(\theta^* , \theta_0) = \frac{1}{\mathcal L^{'}(\theta_0)} .
\end{equation}
Next, we utilize the factors on both source $(\mathcal D_{s(i)}, \mathcal T_{s(i)})$ and target $(\mathcal D_t, \mathcal T_t)$. 

\begin{equation}
\label{eq:factor_source}
\mathcal{f}_{(\mathcal D_{s(i)}, \mathcal T_{s(i)})}(\theta^* , \theta_0) = \frac{1}{\mathcal L_{(\mathcal D_{s(i)}, \mathcal T_{s(i)})}^{'}(\theta_0)}.
\end{equation}

\begin{equation}
\label{eq:factor_target}
\mathcal{f}_{(\mathcal D_t, \mathcal T_t)}(\theta^* , \theta_0) = \frac{1}{\mathcal L_{(\mathcal D_t, \mathcal T_t)}^{'}(\theta_0)}.
\end{equation}

We define the gap between $\theta_{(\mathcal D_{s(i)}, \mathcal T_{s(i)})}^*$ and $\theta_{(\mathcal D_t, \mathcal T_t)}^*$ as $\mathcal{g}(\theta_{(\mathcal D_{s(i)}, \mathcal T_{s(i)})}^*, \theta_{(\mathcal D_t, \mathcal T_t)}^*)$, where $\theta_{(\mathcal D_{s(i)}, \mathcal T_{s(i)})}^*$ and $\theta_{(\mathcal D_t, \mathcal T_t)}^*$ represents the optimal model parameters on the source and target, respectively. Since the value of the factor could be positive or negative, the absolute value of the two factors' subtraction is used to represent the gap between $\mathcal{f}_{(\mathcal D_{s(i)}, \mathcal T_{s(i)})}(\theta^*, \theta_0)$ and $\mathcal{f}_{(\mathcal D_t, \mathcal T_t)}(\theta^*, \theta_0)$. 


\begin{small}
\begin{align}
\label{eq:dist_source_target}
&\mathcal{g}(\theta_{(\mathcal D_{s(i)}, \mathcal T_{s(i)})}^*, \theta_{(\mathcal D_t, \mathcal T_t)}^*)
\notag
\\& 
\notag
\\=\quad& \lvert \mathcal{f}_{(\mathcal D_{s(i)}, \mathcal T_{s(i)})}(\theta^*, \theta_0) - \mathcal{f}_{(\mathcal D_t, \mathcal T_t)}(\theta^* , \theta_0) \rvert
\notag
\\=\quad& \lvert \frac{1}{\mathcal L_{(\mathcal D_{s(i)}, \mathcal T_{s(i)})}^{'}(\theta_0)} - \frac{1}{\mathcal L_{(\mathcal D_t, \mathcal T_t)}^{'}(\theta_0)} \rvert
\notag
\\=\quad& \lvert \frac{\mathcal L_{(\mathcal D_t, \mathcal T_t)}^{'}(\theta_0) - \mathcal L_{(\mathcal D_{s(i)}, \mathcal T_{s(i)})}^{'}(\theta_0)}{\mathcal L_{(\mathcal D_t, \mathcal T_t)}^{'}(\theta_0)\mathcal L_{(\mathcal D_{s(i)}, \mathcal T_{s(i)})}^{'}(\theta_0)} \rvert
\notag
\\=\quad& \frac{\lvert \mathcal L_{(\mathcal D_t, \mathcal T_t)}^{'}(\theta_0) - \mathcal L_{(\mathcal D_{s(i)}, \mathcal T_{s(i)})}^{'}(\theta_0) \rvert}{\lvert \mathcal L_{(\mathcal D_t, \mathcal T_t)}^{'}(\theta_0)\mathcal L_{(\mathcal D_{s(i)}, \mathcal T_{s(i)})}^{'}(\theta_0) \rvert}.
\end{align}
\end{small}

In practice, a model has numerous parameters, thus we expand the above derivation to a high dimension version $\mathcal G'[(\mathcal D_{s(i)}, \mathcal T_{s(i)});(\mathcal D_t, \mathcal T_t)]$. 
\begin{small}
\begin{align}
\label{eq:score}
&\mathcal G'[(\mathcal D_{s(i)}, \mathcal T_{s(i)});(\mathcal D_t, \mathcal T_t)]
\notag
\\=\quad& \parallel\mathcal{F}_{(\mathcal D_{s(i)}, \mathcal T_{s(i)})}(\theta^* , \theta_0) - \mathcal{F}_{(\mathcal D_t, \mathcal T_t)}(\theta^* , \theta_0)\parallel _2
\notag
\\=\quad& \frac{\parallel\nabla \mathcal L_{(\mathcal D_t, \mathcal T_t)}(\theta_0)-\nabla \mathcal L_{(\mathcal D_{s(i)}, \mathcal T_{s(i)})}(\theta_0)\parallel _2}{\parallel \nabla \mathcal L_{(\mathcal D_t, \mathcal T_t)}(\theta_0)\enspace\nabla \mathcal L_{(\mathcal D_{s(i)}, \mathcal T_{s(i)})}(\theta_0)\parallel_2 }.
\end{align}
\end{small}
Similar with Eq.~\ref{eq:factor}, $\mathcal{F}_{(\mathcal D_{s(i)}, \mathcal T_{s(i)})}(\theta^* , \theta_0)$ and $\mathcal{F}_{(\mathcal D_t, \mathcal T_t)}(\theta^* , \theta_0)$ are two vectors, measuring the optimization difficulty from $\theta_0$ to $\theta_{(\mathcal D_{s(i)}, \mathcal T_{s(i)})}^*$ and $\theta_{(\mathcal D_t, \mathcal T_t)}^*$, respectively. 
We then deflate $\frac{\parallel\nabla \mathcal L_{(\mathcal D_t, \mathcal T_t)}(\theta_0)-\nabla \mathcal L_{(\mathcal D_{s(i)}, \mathcal T_{s(i)})}(\theta_0)\parallel _2}{\parallel \nabla \mathcal L_{(\mathcal D_t, \mathcal T_t)}(\theta_0)\enspace\nabla \mathcal L_{(\mathcal D_{s(i)}, \mathcal T_{s(i)})}(\theta_0)\parallel_2 }$ with Schwarz inequality~\cite{steele2004cauchy}.

\begin{equation}
\frac{\parallel\nabla \mathcal L_{(\mathcal D_t, \mathcal T_t)}(\theta_0)-\nabla \mathcal L_{(\mathcal D_{s(i)}, \mathcal T_{s(i)})}(\theta_0)\parallel _2}{\parallel \nabla \mathcal L_{(\mathcal D_t, \mathcal T_t)}(\theta_0)\enspace\nabla \mathcal L_{(\mathcal D_{s(i)}, \mathcal T_{s(i)})}(\theta_0)\parallel_2 }\quad\ge \nonumber
\end{equation}
\begin{equation}
\\\quad\quad\quad\quad\quad\quad\frac{\parallel\nabla \mathcal L_{(\mathcal D_t, \mathcal T_t)}(\theta_0)-\nabla \mathcal L_{(\mathcal D_{s(i)}, \mathcal T_{s(i)})}(\theta_0)\parallel _2}{\parallel \nabla \mathcal L_{(\mathcal D_t, \mathcal T_t)}(\theta_0)\parallel_2\enspace\parallel\nabla \mathcal L_{(\mathcal D_{s(i)}, \mathcal T_{s(i)})}(\theta_0)\parallel_2 }.
\end{equation}
Considering calculating the gradient once might obtain abnormal gradient, we employ PGE (Sec.~\ref{sec:fast_approx}) to calculate the transferability gap which is defined as $\mathcal G[(\mathcal D_{s(i)}, \mathcal T_{s(i)});(\mathcal D_t, \mathcal T_t)]$.

\begin{small}
\begin{align}
\label{eq:calu_scores}
&\mathcal G[(\mathcal D_{s(i)}, \mathcal T_{s(i)});(\mathcal D_t, \mathcal T_t)] = 
\frac{\parallel PGE_{(\mathcal D_t, \mathcal T_t)} - PGE_{(\mathcal D_{s(i)}, \mathcal T_{s(i)})}\parallel _2}{\parallel PGE_{(\mathcal D_t, \mathcal T_t)} \parallel _2\enspace \parallel PGE_{(\mathcal D_{s(i)}, \mathcal T_{s(i)})} \parallel_2} 
\notag
\\& = \frac{\parallel\mathbb E_{\theta_0}\left[  \nabla \mathcal L_{(\mathcal D_t, \mathcal T_t)}( \theta_0 )\right ]-\mathbb E_{\theta_0}\left [ \nabla \mathcal L_{(\mathcal D_{s(i)}, \mathcal T_{s(i)})}( \theta_0 ) \right ] \parallel _2}{\parallel \mathbb E_{\theta_0}\left [ \nabla \mathcal L_{(\mathcal D_t, \mathcal T_t)}( \theta_0 ) \right ] \parallel _2\enspace \parallel \mathbb E_{\theta_0}\left [ \nabla \mathcal L_{(\mathcal D_{s(i)}, \mathcal T_{s(i)})}( \theta_0 ) \right ] \parallel_2}.
\end{align}
\end{small}
It should be noted that only the gradient of the model's backbone $w$ is applied when calculating $\mathcal G[(\mathcal D_{s(i)}, \mathcal T_{s(i)});(\mathcal D_t, \mathcal T_t)]$. Essentially, this metric measures the disparity between $(\mathcal D_{s(i)}, \mathcal T_{s(i)})$ and $(\mathcal D_t, \mathcal T_t)$. 
Finally, we calculate the transferability scores between each $(\mathcal D_{s(i)}, \mathcal T_{s(i)})$ in $\{(\mathcal D_{s(i)}, \mathcal T_{s(i)})\}_{i=1}^N$ and $(\mathcal D_t, \mathcal T_t)$ with Eq.~\ref{eq:calu_scores}.
The algorithm for acquiring the transferability score with PGE is outlined in Algorithm~\ref{alg:PGE}.
\renewcommand{\thealgorithm}{1} 
\begin{algorithm}
    \caption{Principal Gradient Expectation} 
    \begin{algorithmic} 
        \Require A model $\boldsymbol{\mathit M}$ with random initialization $\theta_0$, several different source $\{(\mathcal D_{s(i)}, \mathcal T_{s(i)})\}_{i=1}^N$, and a target $(\mathcal D_t, \mathcal T_t)$.
        \Ensure Ranking of the results of transfer of all sources to the target with $\boldsymbol{\mathit M}$.
        
        
        \For{$n = 1 \to N+1$} \quad\quad\quad//$N+1$ denotes N sources and a target.
            \For{$i = 1 \to I$}  \quad\quad\quad//$I$ denotes the number of iterations.
                \State Input some instances into the model $\boldsymbol{\mathit M}$.
                \State Computing $\nabla \mathcal L( \theta_0)$
                \State $\mathbb E_{\theta_0}[\nabla \mathcal L(\theta_0)]  \leftarrow ((i-1)\ast \mathbb E_{\theta_0}[\nabla \mathcal L(\theta_0)]  + \nabla \mathcal L(\theta_0)) / i$ \quad\quad\quad//The gradients obtained this time are added to the previously collected gradients and averaged. 
            \EndFor
        \EndFor
        \For{$i = 1 \to N$} \quad\quad\quad//$N$ denotes the number of sources
                \State Calculate $\frac{\parallel\mathbb E_{\theta_0}\left[  \nabla \mathcal L_{(\mathcal D_t, \mathcal T_t)}( \theta_0 )\right ]-\mathbb E_{\theta_0}\left [ \nabla \mathcal L_{(\mathcal D_{s(i)}, \mathcal T_{s(i)})}( \theta_0 ) \right ] \parallel _2}{\parallel \mathbb E_{\theta_0}\left [ \nabla \mathcal L_{(\mathcal D_t, \mathcal T_t)}( \theta_0 ) \right ] \parallel _2\enspace \parallel \mathbb E_{\theta_0}\left [ \nabla \mathcal L_{(\mathcal D_{s(i)}, \mathcal T_{s(i)})}( \theta_0 ) \right ] \parallel_2}$
        \EndFor
        \State Sort
        \State \Return Ranking
    \end{algorithmic}
\label{alg:PGE}
\end{algorithm}

\subsection{Robust Evaluation with Multiple Subsampling}\label{sec:robust_eva}
In this section, we introduce the standardized evaluation for the transferability between different domains.
\par
We define $\mathcal A$ as the performance on the target transferred from the pre-trained model. If $\mathcal A((\mathcal D_{s(i)}, \mathcal T_{s(i)});(\mathcal D_t, \mathcal T_t))> \mathcal A((\mathcal D_{s(j)}, \mathcal T_{s(j)});(\mathcal D_t, \mathcal T_t)))$, we expect $\mathcal G[(\mathcal D_{s(i)}, \mathcal T_{s(i)});(\mathcal D_t, \mathcal T_t)] < \mathcal G[(\mathcal D_{s(j)}, \mathcal T_{s(j)});(\mathcal D_t, \mathcal T_t)]$. 
The reliability of our method is validated by computing the correlations between the ranking of $\{\mathcal G[(\mathcal D_{s(i)}, \mathcal T_{s(i)});(\mathcal D_t, \mathcal T_t)]\}_{i=1}^N$ and the ranking of $\{\mathcal A((\mathcal D_{s(i)}, \mathcal T_{s(i)});(\mathcal D_t, \mathcal T_t))\}_{i=1}^N$. Kendall's $\tau$~\cite{fagin2003comparing} coefficient, a measure of rank correlation, quantifies the similarity between two rankings. Kendall's $\tau$ coefficient is defined as:
\begin{equation}
\label{eq:Kendall}
{\displaystyle \tau ={\frac {2}{n(n-1)}}\sum _{i<j}\operatorname {sgn}(x_{i}-x_{j})\operatorname {sgn}(y_{i}-y_{j})}.
\end{equation}
$x_i \in \mathcal{X}$ and $y_i \in \mathcal{Y}$ ($\mathcal{X}$ and $\mathcal{Y}$ represent two rankings respectively). And n indicates the number of items in the ranking. The $sgn(\cdot)$ function is a symbolic function.
Specifically, the range of $\tau$ is $\left [ -1,1\right ]$, a higher $\tau$ indicates a stronger correlation between $\{\mathcal G[(\mathcal D_{s(i)}, \mathcal T_{s(i)});(\mathcal D_t, \mathcal T_t)]\}_{i=1}^N$ and $\{\mathcal A((\mathcal D_{s(i)}, \mathcal T_{s(i)});(\mathcal D_t, \mathcal T_t))\}_{i=1}^N$. $\tau = 0 $ shows no correlation between them. 

Following~\cite{nguyen2020leep, pandy2022transferability}, we generate multiple subsets of the target domain $\mathcal D_t$ with two approaches:  
$\emph{I}_1$ randomly selects $\eta_1\%$ categories in $\mathcal D_t$ and all samples in the selected categories are used.
$\emph{I}_2$ randomly selects $\eta_2\%$ samples from each category in $\mathcal D_t$. 
Concretely, we construct one hundred different $(\mathcal D_t, \mathcal T_t)$ from $\mathcal D_t$ with the two approaches mentioned above and evaluate the performance of the pre-trained model with the selected target test set. A curve is drawn in which the horizontal axis represents the sizes of the subsets and the vertical axis shows the performance of the different test sets.
 We use the area under the curve as the measurement of the $\mathcal A((\mathcal D_{s(i)}, \mathcal T_{s(i)});(\mathcal D_t, \mathcal T_t))$. 
\par



\begin{figure*}[htbp]
    	\centering
    	\begin{subfigure}{1\linewidth}
    		\centering
    		\includegraphics[width=1\linewidth]{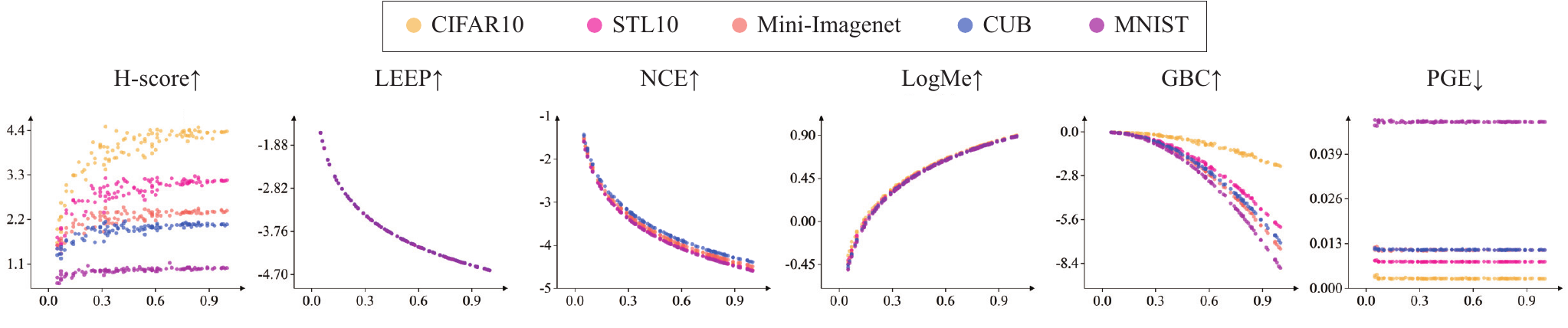}
    		\caption{The subsets are constructed by \textbf{SI}}
    		\label{stability_SI}
    	\end{subfigure}
    	\centering
    	\begin{subfigure}{1\linewidth}
    		\centering
    		\includegraphics[width=1\linewidth]{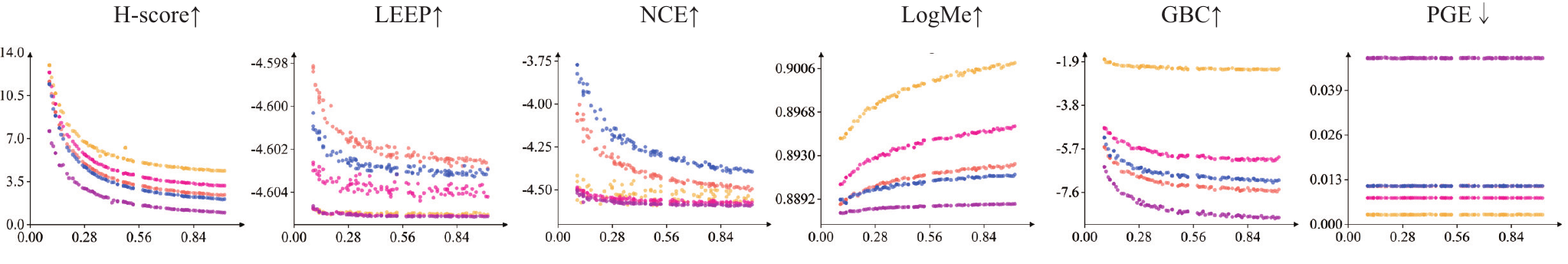}
    		\caption{The subsets are constructed by \textbf{SII}}
    		\label{stability_SII}
    	\end{subfigure}
    	\caption{
     The stability comparison of different methods. $\uparrow$ denotes that a higher value represents a better source, and $\downarrow$ denotes that a lower value indicates a better source. We show the transferability scores' variation of different measurement techniques along with increasing sampling ratios of subsets.
     The top row of the figure illustrates that we build target subsets with strategy \textbf{SI} on CIFAR100. We compare the results with five existing transferability metrics. In each plot, the horizontal axis means the sampling ratio of each subset while the vertical axis is the transferability score.
    	The bottom row of the figure indicates that we construct subsets with strategy \textbf{SII} on CIFAR100.}
    	\label{fig:stability}
\end{figure*}

\section{Experiment}\label{sec:exp}
In this section, we scrutinize and affirm the proposed transferability metric with diverse image data. Initially, we introduce the experimental settings, followed by an evaluation of the proposed PGE from four key perspectives: stability, reliability, efficiency, and generalizability.

\subsection{Experimental Settings}

\noindent \textbf{Datasets.}
Our experiments are conducted on the following eight datasets: CIFAR10~\cite{krizhevsky2009learning}, CIFAR100~\cite{krizhevsky2009learning}, STL10~\cite{coates2011analysis}, Mini-ImageNet~\cite{vinyals2016matching}, CUB~\cite{wah2011caltech}, MNIST~\cite{deng2012MNIST}, FGVC-Aircraft (Aircraft)~\cite{maji2013fine}, and PASCAL VOC 2012~\cite{everingham2015pascal}. Table~\ref{tab:dataset} provides a summary of the characteristics of these datasets, including resolution size (Resolution), number of images (Img), and number of classes (Class).

\begin{table}[htbp]
\begin{tabular}{ccccc}
\hline
Dataset    & Resolution        & Img  & Class \\ \hline
CIFAR10~\cite{krizhevsky2009learning}    & $32\times 32$  & 60k        & 10      \\
CIFAR100~\cite{krizhevsky2009learning} & $32\times 32$ & 60k          & 100      \\
STL10~\cite{coates2011analysis}    & $96\times 96$  & 13k        & 10       \\
Mini-ImageNet~\cite{vinyals2016matching}    & $84\times 84$  & 60k        & 200       \\
CUB~\cite{wah2011caltech}    & $512\times 512$  & 11.7k        & 200       \\
MNIST~\cite{deng2012MNIST}    & $28\times 28$  & 70k        & 10       \\
Aircraft~\cite{maji2013fine}    & $512\times 512$  & 10k        & 100       \\
PASCAL VOC 2012~\cite{everingham2015pascal}    & $320\times 480$  & 1.7k        & 21       \\
\hline
\end{tabular}
\caption{The summary of datasets.}
\label{tab:dataset}
\end{table}

\noindent \textbf{Transfer Methods.} 
While previous studies have concentrated more on linear probing, we examine both fine-tuning and linear probing.

\noindent(1) \emph{Linear Probing}. This technique involves freezing the feature extractor of the model and training a task-specific layer from scratch with the target dataset.

\noindent(2) \emph{Fine-Tuning}. Here, we replace the model's task-specific layer with a new one and fine-tune the entire model, including the feature extractor and the task-specific layer, on the target dataset.

\noindent \textbf{Evaluation.} 
We construct the subsets with two distinct methods as follows:

\noindent\textbf{SI}: The first approach randomly samples 5\% to 100\% of the target categories and uses all images within these categories.

\noindent\textbf{SII}: The second approach randomly selects a percentage between 10\% to 100\%  images within each category. 

Notably, we do not sample all target datasets with either of the two sampling approaches. And we generate subsets with~\textbf{SI} when the number of samples in each category is relatively small.

\noindent \textbf{Implementation Details}.
We run 600 epochs of linear probing and fine-tuning for each subset of the target domain (using SGD without Momentum and Cosine annealing)~\cite{loshchilov2016sgdr}. And we adjust the learning rate with an initial one as $0.1\ast batch size / 256$. 
Three classic backbones are adopted in the experiments, i.e., Resnet18~\cite{he2016deep}, Resnet50~\cite{he2016deep}, and VGG16~\cite{simonyan2014very}. Additional experiments with more datasets and large-scale models are shown in Appendix.

\subsection{Stability Comparison}
In this section, we compare the proposed PGE with existing techniques, including LEEP~\cite{nguyen2020leep}, LogMe~\cite{you2021logme}, H-score~\cite{bao2019information}, NCE~\cite{tran2019transferability}, and GBC~\cite{pandy2022transferability}. Moreover, we use CIFAR10, STL10, Mini-ImageNet, CUB, and MNIST as source datasets and CIFAR100 as the target dataset. 






Figure~\ref{fig:stability} shows the variation of transfer scores among different measurement techniques when increasing the sampling ratio of each subset. It is acknowledged that even with different subsets of a target domain, the measurement results should be stable.
In Figure~\ref{fig:stability}(a), the subsets are constructed by \textbf{SI}.
We observed that as the sampling ratio arises, H-score~\cite{bao2019information} and LogMe~\cite{you2021logme} increase while LEEP, NCE, and GBC decrease. By contrast, our proposed PGE is superior to all compared methods for stability, as the sampling ratio has no effect on its outputs. So we argue that the distribution of the target domain is an important factor to compute the transferability for the compared methods, as the scores vary with different distributions. Since our calculation does not involve the distribution, it is more stable than others. Moreover, the curves of  LEEP~\cite{nguyen2020leep}, LogMe~\cite{you2021logme}, and NCE~\cite{tran2019transferability} almost overlap when randomly sampling categories over CIFAR100 as \textbf{SI}, which demonstrate that it's hard to discriminate which is the best source for the target because they tend to yield similar transferability scores. In addition, although H-score and GBC gain the same correct source (i.e., CIFAR 10) as our PGE, PGE produces more stable and distinguishable results. And we can get the correct result with only a portion of categories from the target dataset.
In Figure~\ref{fig:stability}(b), the subsets are constructed by~\textbf{SII}. A similar conclusion could be drawn as in Figure~\ref{fig:stability}(a) that PGE shows remarkable advantages over other methods. We think these approaches may fail when the number of images is small since the data distribution of a small target dataset usually cannot represent the true distribution of real-world data. In comparison, PGE uses the expectation of the principle gradient to estimate the transferability gap, as the expectation can effectively reduce the impact of abnormal gradients. Therefore, the proposed PGE can still obtain the correct result. 
In summary, the proposed PGE yields more consistent and distinguishable results for both strategies. 

\begin{table*}[!t]
\centering
\begin{tabular}{cccccccccccc}
\Xhline{1 pt}
& \multicolumn{2}{c}{CIFAR10~\cite{krizhevsky2009learning}} & \multicolumn{2}{c}{STL10~\cite{coates2011analysis}} & \multicolumn{2}{c}{CIFAR100~\cite{krizhevsky2009learning}} & \multicolumn{2}{c}{CUB~\cite{wah2011caltech}} & \multicolumn{2}{c}{Aircraft~\cite{maji2013fine}} & \multirow{2}*{Average Kendall's $\tau$} \\
& LP           & FT           & LP          & FT          & LP           & FT            & LP         & FT         & LP            & FT           &                          \\
\Xhline{0.5 pt}
LEEP)~\cite{nguyen2020leep}  & 0.19         & 0            & 0           & 0           & 0            & -0.6          & 0.19       & 0.2        & -0.19         & -0.19        & -0.04                    \\
H-score)~\cite{bao2019information}     & \textbf{1}            & \textbf{0.79}         & \textbf{1}           & \textbf{1}           & \textbf{1}            & 0             & -0.19      & 0.19       & 0.19          & 0.19         & 0.52                    \\
NCE)~\cite{tran2019transferability}   & 0.39         & 0.19         & -0.2        & -0.2        & 0            & -0.6          & -0.19      & -0.39      & -0.79         & -0.79        & -0.26                   \\
LogMe)~\cite{you2021logme} & \textbf{1}            & \textbf{0.79}          & \textbf{1}           & \textbf{1}           & \textbf{1}            & 0             & -0.39      & -0.4       & -0.79         & -0.4         & 0.281                    \\
GBC)~\cite{pandy2022transferability}   & \textbf{1}            & \textbf{0.79}          & 0.4         & 0.4         & \textbf{1}            & 0             & 0.39       & 0.39       & 0.19          & -0.19        & 0.44         \\
\Xhline{0.5 pt}
\textbf{PGE} (Ours) & \textbf{1}            & \textbf{0.79}          & \textbf{1}         & \textbf{1}         & \textbf{1}            & 0             & \textbf{0.39}        & \textbf{0.39}        & \textbf{0.79}          & \textbf{0.39}          & \textbf{0.68}\\
\Xhline{1 pt}
\end{tabular}
\caption{The reliability comparison of different methods. The value (higher is better) indicates the correlation between the ranking calculated by the transferability estimation technique and the ranking of transfer performances. Both the source and target tasks are image classification. And we describe the results for five target domains with two transfer methods (Linear Probing (LP) and Fine-Tuning (FT)). It could be found that the proposed method obtains the highest Kendall's $\tau$)~\cite{fagin2003comparing} coefficient in most experiments and the highest average Kendall's $\tau$~\cite{fagin2003comparing} coefficient for the different domains.}
\label{tab:reliability_comparison_cls}
\end{table*}

\begin{table}[!t]
\centering 
\begin{tabular}{ccccllll}
\Xhline{1 pt}
\textbf{}      & \textbf{PGE Gap $\downarrow$} ($10^{-2}$) & \textbf{LP(\%)} & \textbf{FT(\%)}\\
\Xhline{0.5 pt}
CIFAR100       & 0.284               & 50.57               & 66.69\\
STL10          & 0.773               & 39.86               & 65.39\\
Mini-ImageNet & 1.118               & 32.04               & 65.28\\
CUB            & 1.125               & 26.20               & 60.67\\
MNIST          & 4.842               & 17.32               & 62.66\\
\Xhline{0.5 pt}
               &                      & $\tau$\;:\;1  & $\tau$\;:\;0.79\\
\Xhline{1 pt}
\end{tabular}
\caption{
The reliability of PGE scores. $\downarrow$ denotes that a lower gap which represents a better source. Transfer performances were obtained with two transfer methods, i.e., Linear Probing (LP) and Fine-Tuning (FT).
}
\label{tab:PGE_transfer_results_cls}  
\end{table}

\begin{table}[!t]
\begin{tabular}{ccclllll}
\Xhline{1 pt}
\textbf{}    & \textbf{PGE Gap $\downarrow$} ($10^{-2}$) & \textbf{MIoU(\%)}\\
\Xhline{0.5 pt}
CIFAR100     & 3.86            & 65.09\\
CIFAR10      & 4.04            & 65.03\\
Mini-ImageNet & 1.17            & 67.85\\
STL10        & 1.23            & 66.07\\
CUB          & 1.18            & 65.53\\
\Xhline{0.5 pt}
             & \multicolumn{2}{c}{$\tau$\;:\;0.79}\\
\Xhline{1 pt}
\end{tabular}
\caption{The generalizability of PGE with PASCAL VOC 2012 dataset. $\downarrow$ denotes that a lower gap which represents a better source.}
\label{tab:generalizability_cls_to_seg}  
\end{table}

\subsection{Reliability Comparison}
In this section, we adopt the consistency between the estimated results and the real transfer performances to compare the reliability of different methods. As mentioned in Section~\ref{sec:method}, the transferability is supposed to correlate well with the final performance of a model after fine-tuning/linear probing on the target task. To validate the reliability of the proposed PGE, we conduct experiments with both fine-tuning and linear probing on CIFAR10, CIFAR100, STL10, CUB, and Aircraft. We choose five different sources for each target. As shown in Table~\ref{tab:reliability_comparison_cls}, the relevance between the transfer performances and transferability is computed with Kendall's $\tau$~\cite{fagin2003comparing} coefficient. 
In the linear probing  process, PGE identifies the optimal source for all five targets. In the fine-tuning process, PGE determines the optimal source for four targets. Moreover, we achieve the highest correlation for CIFAR10, CUB, and Aircraft with both fine-tuning and linear probing. 
Overall, the proposed PGE obtains the highest average correlation of $0.68$ among all existing approaches. 

The results in Table~\ref{tab:reliability_comparison_cls} reveal that H-score and LogMe can produce competitive results on simple datasets (CIFAR10, STL10, and CIFAR100). However, most of the existing approaches show poor performance on more challenging datasets (CUB and Aircraft). We consider that existing methods are sensitive to the resolution of the images while the proposed PGE is robust enough to alleviate this issue.


Specifically, Table~\ref{tab:PGE_transfer_results_cls} presents the results of the estimated transferability and the transfer performance on CIFAR10. When linear probing is adopted, the transfer performances completely match ($\tau = 1 $) the ranking of the PGE results. For fine-tuning, we also obtain Kendall's $\tau = 0.79$, illustrating the transfer performances and PGE results are well-aligned. 
\par

\begin{table*}[!t]
\centering 
\resizebox{\textwidth}{!}{
\begin{tabular}{cccccccc}
\Xhline{1.5 pt}
\textbf{Pre-trained task} & \textbf{Transfer Target} & \textbf{H-score~\cite{bao2019information}} & \textbf{ LEEP~\cite{nguyen2020leep}} & \textbf{NCE~\cite{tran2019transferability}} & \textbf{GBC~\cite{pandy2022transferability}} & \textbf{LogMe~\cite{you2021logme}} & \textbf{PGE (ours)} \\
\Xhline{0.5 pt}
Classification     & Classification  & \ding{51}                & \ding{51}             & \ding{51}            & \ding{51}            & \ding{51}              & \ding{51}            \\
Classification     & Regression      & \ding{55}                & \ding{55}             & \ding{55}            & \ding{55}            & \ding{51}              & \ding{51}            \\
Unsupervised        & Classification  & \ding{55}                & \ding{55}             & \ding{55}            & \ding{55}            & \ding{51}              & \ding{51}            \\
Unsupervised        & Regression      & \ding{55}                & \ding{55}             & \ding{55}            & \ding{55}            & \ding{51}              & \ding{51}\\
\Xhline{1.5 pt}
\end{tabular}
}
\caption{Generalizability of different techniques for different transfer settings. 
``\ding{51} / \ding{55}" indicates whether the technique is applicable/unapplicable for this transfer setting. }
\label{tab:generalizability_comparison} 
\end{table*}

\begin{table}[!t]
\centering
\resizebox{0.48\textwidth}{!}{
\begin{tabular}{ccccc}
\Xhline{1 pt}
              & PGE Gap $\downarrow$ (\textbf{Sup.}) & PGE Gap $\downarrow$ (\textbf{ Uns. ($10^{-2}$)}) & \textbf{LP (\%)} & \textbf{FT (\%)} \\
\Xhline{0.5 pt}
Mini-ImageNet & 4.35             & 5.00             & 48.21      & 61.83      \\
CIFAR10       & 1.91             & 5.80             & 45.43      & 61.36      \\
CIFAR100      & 2.56             & 7.79             & 45.34      & 59.47      \\
CUB           & 4.75             & 7.81             & 37.93      & 55.82      \\
\Xhline{1 pt}
\end{tabular}}
\caption{The ablation study with supervised (\textbf{Sup.}) and unsupervised 
 (\textbf{Uns.}) strategies of obtaining gradients. $\downarrow$ denotes that a lower gap which represents a better source.}
\label{tab:ablation_sup_unsup}  
\end{table}

\subsection{Efficiency Comparison}
Regarding a new source, the existing methods need the pre-trained parameters trained on the new domain to measure the transferability, which is time-consuming.
In contrast, our proposed method is more computationally efficient for the following reasons. First, by employing the gradient of the first-order optimization, our method does not require any pre-training process for the source. 
Second, by analyzing the stability of the proposed method above, we can find that the proposed PGE can significantly reduce the computational cost by reducing the amount of data, which is not applicable to all other methods. Table~\ref{time} compares the efficiency. We speed up the estimation process by up to $7\times$ faster than existing works while obtaining more accurate transferability.
\begin{table}[htbp]
\centering 
\resizebox{0.5\textwidth}{!}{
\begin{tabular}{ccccccc}
\Xhline{1.5 pt}
\textbf{Method} &  H-score & LEEP & NCE & GBC & LogMe & PGE (ours) \\
\Xhline{0.5 pt}
\textbf{Time(s)}        & 327159      & 327135                & 327131             & 327212            & 327299            & 45488\\
\Xhline{1.5 pt}
\end{tabular}
}
\caption{The time burden of transferability estimation from CIFAR10 to CIFAR100 with ResNet18.}
\label{time}  
\vspace{-0.5cm}
\end{table}

\subsection{Generalizability Comparison}
We further extend our proposed PGE to compute transferability between source and target tasks when they are different. The source domains are CIFAR10, CIFAR100, STL10, Mini-ImageNet, and CUB while the target domain is PASCAL VOC 2012. The estimation process aims to find the most suitable source domain for the segmentation task based on those classification tasks. The backbone here is VGG16 and the epochs for pre-training is 100. The results are shown in Table~\ref{tab:generalizability_cls_to_seg}. The PGE values in Table~\ref{tab:generalizability_cls_to_seg} suggest that Mini-ImageNet could be a suitable source, which is verified by the segmentation accuracy. 


We list four common transfer settings, as shown in Table~\ref{tab:generalizability_comparison}. It could be concluded that all of the approaches can be easily adapted to classification tasks. However, when the source task and target task are different, H-score, LEEP, NCE and GBC can not estimate transferability. By contrast, our proposed PGE and LogMe are more practical to adapt to this setting. For classification and regression tasks, LogMe designs different modules while our proposed PGE computes the transferability uniformly without pre-training.


\subsection{Ablation Study}
Recent researches ~\cite{zhao2020makes,islam2021broad} has shown that unsupervised techniques acquire more low-level and mid-level information. The information is more readily adaptable to a new domain than supervised techniques. In supervised learning processes, models tend to learn high-level semantics. In this subsection, we examine transferability with principle gradient expectation which is obtained by supervised and unsupervised techniques. The results are shown in Table~\ref{tab:ablation_sup_unsup}. It has been experimentally found that the supervised method is more susceptible than the unsupervised method. A reasonable explanation for this is that category information is embedded into the model to improve discrimination but is harmful to calculating transferability.

\section{Conclusion}
Determining which source is the best for a particular target task is challenging. Moreover, it is computationally costly to determine the source by fine-tuning/linear probing all possible combinations of the sources and target task. In this work, we summarize the properties that a good transferability metric should possess. Building upon them, we propose a simple yet effective transferability estimation approach termed PGE based on principal gradient expectation. To properly evaluate the method's validity, we applied two sub-sampling techniques to the target domain. The experimental results on both fine-tuning and linear probing demonstrate that PGE is superior to existing metrics. Furthermore, since PGE computes the gradient of the backbone under the unsupervised mode, it is more flexible and can be extended to different tasks. In the future, it is worth exploring choosing a proper source task or even a rational composition of source tasks for a given target task.

{\small
\bibliographystyle{ieee_fullname}
\bibliography{egbib}
}
\end{document}